\documentclass[conference,letterpaper,10pt]{IEEEtran}  %

\usepackage{amsmath,amsfonts}
\usepackage[linesnumbered,ruled,vlined]{algorithm2e}
\usepackage{array}
\usepackage{textcomp}
\usepackage{xcolor}
\usepackage{stfloats}
\usepackage{url}
\usepackage{verbatim}
\usepackage{graphicx}
\usepackage{multirow}
\usepackage{cite}
\usepackage{inputenc}
\usepackage{hyperref}
\hypersetup{hidelinks}

\usepackage{bbm}
\usepackage{siunitx}
\usepackage{adjustbox}

\SetKwInput{kwInit}{Init}
\SetKwRepeat{Do}{do}{while}

\usepackage{multirow}
\usepackage{makecell}
\usepackage{booktabs}

\usepackage{amssymb}
\usepackage{dsfont}
\usepackage{float}
\usepackage{subfigure}
\usepackage{caption}
\usepackage{threeparttable}
\usepackage[acronym, nopostdot, style=super, nonumberlist]{glossaries}
\DeclareCaptionLabelFormat{andtable}{#1˜#2 \& \tablename˜\thetable}
\usepackage[normalem]{ulem}
\useunder{\uline}{\ul}{}
\makeglossaries

\newcommand{\ie}{\textit{i}.\textit{e}.,}

\newcommand*{\prob}{\mathsf{P}}

\usepackage{amsthm}
\theoremstyle{definition}

\theoremstyle{plain}

\IEEEoverridecommandlockouts                              %

\begin{document}

\title{\LARGE \bf
Hi-CoLA: High Confidence Lower Bound Approximation Based Reinforcement Learning for Flex-Route Transit Operation Control
}

\author{Qi Kang$^{1}$, Haris N. Koutsopoulos$^{1}$, Hongliang Guo$^{2}$, and Jinhua Zhao$^{3}$\\[2pt]
{\normalsize $^{1}$Northeastern University \qquad
             $^{2}$I2R, A*STAR \qquad
             $^{3}$MIT}}

\maketitle
\thispagestyle{empty}
\pagestyle{empty}

\begin{abstract}

This paper studies safe policy improvement for real-time operation control of flex-route transit (FRT) systems, where robustness is critical and any performance decrease is undesirable. While reinforcement learning (RL) has shown strong empirical success, its adoption remains limited in safety and cost critical applications, such as intelligent transportation, since learned policies lack statistically certified performance guarantees before deployment. In this paper, we propose the High Confidence Lower Bound Approximation (Hi-CoLA) RL framework, which combines importance sampling and concentration inequalities with a differentiable surrogate network to efficiently estimate the high-confidence performance lower bound and optimize it via policy gradient for robust performance improvement. On the FRT operation control problem, Hi-CoLA outperforms state-of-the-art offline and safe RL methods in both average and tail-risk performance. A case study on an actual route in Chicago, implemented in the SUMO simulator, serves as a proof of concept. Hi-CoLA enhances real-time FRT control with guaranteed performance and is broadly applicable to other safety critical decision-making scenarios.

\end{abstract}

\section{Introduction}

\IEEEPARstart{P}{ublic} transit systems worldwide are undergoing a critical transformation in response to increasing urban complexity and the growing demand for sustainable and inclusive mobility solutions \cite{eszes2026drt}. Flex-Route Transit (FRT) has emerged as a promising hybrid solution\cite{Survey1}. In FRT systems, vehicles follow predefined routes and schedules, but are permitted to deviate from their base path to serve on-demand requests within a designated service zone, as shown in Figure 1. 

\begin{figure}[!ht]
    \centering
    \captionsetup{belowskip=0pt}%
        \includegraphics[width=0.48\textwidth]{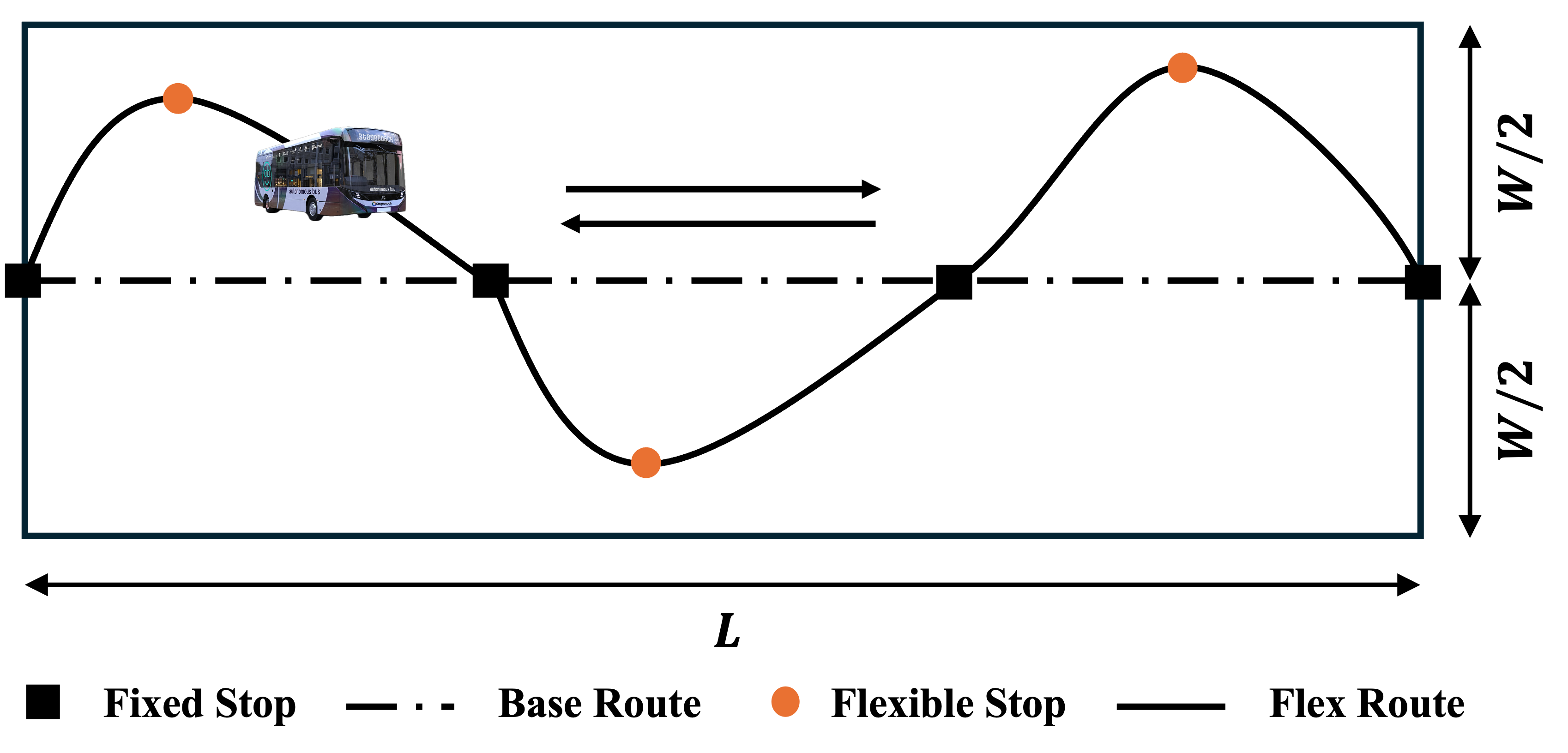}
        \label{fig:1}
    \caption{Illustration of Flex-Route Transit System}

\end{figure}

Historically, FRT performance was limited by manual operations and phone-call-based coordination\cite{randall1996}. Recent advances in autonomous driving and intelligent transportation systems (ITS) \cite{li2026ridepooling,fernando2026its,Shidmoosavi2023smart} facilitate seamless communications and responsive control, transforming FRT from a conceptual model into an operational reality. 

To date, researchers have developed heuristic-based\cite{alshalalfah2026frt} and optimization-based \cite{li2025frtmeeting} approaches for FRT real-time operation control, yet they present notable trade-offs between accuracy and computational efficiency, limiting their applicability for real-time, high-dimensional environments\cite{its2019}. Reinforcement learning (RL) provides a flexible, fast-response solution that adapts to complex, uncertain environments through interaction \cite{qu2025routing,zhang2025uncertainty}, making it well-suited for real-time control of dynamic systems\cite{pan2026learning}. However, applying RL to real-world decision-making is not common\cite{Paduraru2021}, particularly in scenarios involving financial or human safety risks since the learned policies lack statistically certified performance guarantees prior to deployment and exploration in the actual environment can be dangerous. Public transportation exemplifies this challenge, as it demands both cost-efficiency and high operational reliability\cite{wu2024dynamic}. The stochastic nature of RL policies introduces uncertainty to operational decision-making\cite{Paduraru2021}. Moreover, sim-to-real transfer poses a significant barrier; policies trained in simulation often degrade under real-world variability and constraints\cite{hwangbo2019}. For stakeholders, the success of RL in real-world FRT operations can be as uncertain as tossing a coin unless the training process is trustworthy.

This paper aims to bridge the research gap by proposing a high-confidence, performance-guaranteed RL approach for real-time operation control of FRT systems. We introduce a High Confidence Lower Bound Approximation (Hi-CoLA)-based RL training framework that embeds quantified confidence levels directly into the learning process. We first pretrain a Hi-CoLA network with perturbed policies around the behavioral policy and data collected by the behavioral policy to precisely model the relationship between policy parameters and the respective high-confidence performance lower bounds. Using a policy gradient method, we iteratively improve policy performance by maximizing this bound, with parameters updated via backpropagation through the Hi-CoLA network. Training is conducted offline; once the desired performance is achieved, the behavioral policy is updated and deployed in the operational environment to collect new data to enable continual improvement. We verify the performance improvement guarantee of Hi-CoLA in terms of monotonic average return increase and evaluate Hi-CoLA's performance against state-of-the-art offline and safe reinforcement learning approaches in dynamic environments using both average and tail-risk metrics. Then we examine Hi-CoLA's robustness to the quality of the initial behavioral policy. Lastly, we present a high-fidelity traffic simulation-based case study applying Hi-CoLA to a real deviation bus service in Chicago, demonstrating its effectiveness and applicability for real-world decision-making systems.

The contributions of this paper are summarized as follows:
(1) We propose the Hi-CoLA reinforcement learning framework that encodes a quantified confidence level into the training pipeline for real-time operation control. (2) The Hi-CoLA network can be used to estimate the performance lower bound of any given policy with a user-defined confidence level. (3) The proposed framework shows robustness to the quality of the initial behavioral policy and has strong potential in domains with cost sensitivity and challenges in accurate simulation.

\section{Related Work}

In this section, we briefly review related work on the real-time operation control of FRT systems and reinforcement learning approaches, which concern the safety of the policy. For a comprehensive literature review, readers may refer to \cite{Survey1} for FRT systems and \cite{garcia2015,offlinerl} for safe and offline reinforcement learning.
\subsection{Real-time Operations Control for FRT}
State-of-the-practice reports a heavy reliance on simple heuristics originating from static settings where demand and traffic are known in advance. Only a handful of studies consider dynamic operation with real-time decisions. 

\cite{Quadrifoglio2007} first examined dynamic FRT using a schedule-insertion heuristic under a simplified setting with no capacity constraints and rejections, with later extensions to multi-vehicle coordination with dynamic demand \cite{Liu2011} and re-routing with real-time traffic via Google Maps \cite{Qiu2014}. Beyond single-request insertions, \cite{Zhang2023} clusters requests to generate dynamic meeting points and solves a mixed-integer problem with two-stage optimization to reduce rejections. Hybrid optimization–heuristic pipelines have also been explored. For instance, \cite{ZhengLi2019} formulates a mixed-integer linear programming (MILP) for pre-booked demand and then inserts on-the-fly requests, subject to capacity and tolerance on schedule adherence. Furthermore, \cite{ZhengGaoLi2021} extends this approach to complex settings with request cancellations and passengers' no-shows. Overall, these approaches are batched and planning-oriented and thus do not deliver flexible real-time decisions that react to last-minute requests while safeguarding fixed-stop reliability. By contrast, RL learns reactive deviation policies from real-time states and shows consistent gains over heuristic baselines in flex-route operations\cite{joseph2025}.

\subsection{Safe Reinforcement Learning}
The ultimate goal of safe reinforcement learning is to make RL dependable enough for deployment in cost and safety critical settings; however, because “safety” has multiple formal meanings in the literature, research has diverged into distinct paths.
One widely adopted definition, summarized in the comprehensive survey \cite{garcia2015}, frames it as the process of learning policies that maximize the expected return while ensuring reasonable system performance and/or respecting safety constraints during both the learning and deployment phases. This view often uses constrained Markov Decision Process (MDP) formulations, with methods such as Constrained Policy Optimization (CPO) \cite{Achiam2017CPO}, Lyapunov-based RL \cite{Chow2018Lyapunov}, shielding \cite{Alshiekh2018Shielding}, and reachability-based safe exploration\cite{Sui2015SafeOpt}. Risk-sensitive RL, which optimizes tail-aware criteria \cite{Chow2015CVaR} (e.g., conditional value at risk (CVaR) and coherent risks), and robust MDPs\cite{Nilim2005RobustMDP}, which optimize worst-case value over ambiguity sets for the dynamics, also fall in this category. Another perspective defines safety as robustness to arbitrary off-policy sampling. For instance, Retrace$(\lambda)$ \cite{Munos2016Retrace} remains stable and low-variance even when the behavioral and target policies differ substantially, making it suitable for heterogeneous or logged datasets. Closely related, offline reinforcement learning, such as IQL\cite{2022IQL} and CQL\cite{Kumar2020CQL}, focuses on learning from fixed datasets without interaction risks, but distributional shift and unreliable off-policy evaluation pose central challenges.

The definition of safety adopted in this work is from \cite{Thomas2015HCOPE}. Here, safety is treated as a guarantee against regression: given a baseline $\rho\_$ and tolerance $\delta$, the algorithm should return a policy whose expected performance falls below $\rho\_$ with probability of at most $\delta$. Methods such as High-Confidence Policy Improvement (HCPI) \cite{Thomas2015HighConfidencePI} enforce this by combining lower confidence bounds \cite{Thomas2015HCOPE} with conservative updates, ensuring no performance regression at deployment. Beyond these, related progress in offline RL strengthens such guarantees via conservatism. Notably, \cite{Kumar2020CQL} provides lower-bound value guarantees and monotonic improvement tendencies from static datasets; theoretical work shows that pessimism over Bellman-consistent function classes\cite{Xie2021BCPessimism} can be provably efficient under realistic coverage assumptions.

Prior safe RL work either enforces constraints without certifying performance at deployment or targets off-policy stability without guarantees. Even the few methods that do provide explicit statistical guarantees can be slow and data-hungry to realize improvements. We introduce Hi-CoLA, an offline framework that learns a high-confidence lower bound on policy performance from real-world operations data and directly maximizes that performance lower bound, yielding a deploy-or-not certificate against a baseline. This closes the deployment gap while avoiding simulator dependence and enables a simple and automatic cycle of log, improve, and deploy for FRT services.

\section{Problem Formulation}
In this section, we present the formulation of the Flex-Route Transit real-time operation control problem.
We first introduce the system description and then formulate it under the partially observed Markov decision process (POMDP) framework so that we can solve it from a reinforcement learning perspective. Table I presents a list of major notations used throughout the paper. We adopt the convention of using bold symbols to represent a matrix or a vector and non-bold symbols to represent a scalar or a vector element.
\begin{table}[!ht]
\centering
\caption{List of Major Notations Used in the Paper}
\label{tab2:parameters-notations}
\begin{tabular}{l|l} 
\hline
Notations                                                         & Descriptions                                     \\ 
\hline
$W \times L$                          & dimensions of service area     \\
$\mathcal{G}(\mathcal{V},\mathcal{E})$                          & the routing network (represented as a graph)     \\
$\mathcal{V}$                                                     & set of total stops, $|\mathcal{V}|=N$                     \\
$\mathcal{C}$                                                     & set of fixed stops, $|\mathcal{C}|=n$                     \\
$\mathcal{F}$                                                     & set of flexible stops, $|\mathcal{F}|=m$                     \\
$\mathcal{E}$                                                     & set of directed edges connecting stops                     \\
$\mu_T,\sigma_T$                                                     & travel time distribution parameters\\
$\boldsymbol{T}$                                                     & travel time matrix\\
\hline
$K$                                                               & $\#$ of vehicles                \\
$Q_c$                                                     & vehicle capacity                     \\
$h$                                   & departure interval          \\
$W_\text{fixed},W_\text{flex}$                                        & waiting time threshold at fixed and flexible stops                          \\
$\beta_1$                                                            & passenger no-show rate 
\\
$\boldsymbol{Q_t^p}$                                                         & set of passengers waiting at each stop at time t\\
$Q_t^o$                                                               & network occupancy matrix   \\
\hline
                                
$\boldsymbol{\theta}$                                                             & policy parameters                                          \\
$H$                                                        & trajectory length     \\
$c_p$                                                            & size of perturbation space              \\
$\beta_{kl}$                                                        &  KL-divergence threshold    \\

\hline
\end{tabular}

\end{table}                                                                      
\subsection{Flex-Route Transit Operation System Description}
The FRT operation is formulated as a Vehicle Routing Problem with Pick-up and Drop-off (VRPPD). The objective is to serve passenger requests at both fixed and flexible stops within a designated service area by dynamically dispatching and coordinating a fleet of vehicles.
The service area is modeled as a rectangle of dimensions $W \times L$, bounded by two terminal checkpoints. The routing network is represented by a directed graph $\mathcal{G}(\mathcal{V}, \mathcal{E})$, where $\mathcal{V} = \{v_1,v_2,\ldots,v_N\}$ is the set of all the potential stops, and  $\mathcal{E}$ is the set of edges connecting stops. Fixed stops along the base route, which are mandatory to be visited, are denoted as  $\mathcal{C} = \{c_1,c_2,\ldots,c_n\}$ with terminals $c_1$ and $c_n$; flexible stops located between consecutive fixed stops are denoted as $\mathcal{F}= \{f_1, f_2, \ldots,f_{m}\}$. Then $\mathcal{V}$ can be expressed by $\mathcal{V} = \mathcal{C}\cup\mathcal{F}$ and $v_i = c_i$ and $v_{i+n} = f_i $. The directed edge from node $v_i$ to node $v_j$ is denoted as $\{e_{i,j} = (v_i, v_j)|~\forall v_i, v_j \in \mathcal{V},~ i \neq j\}$ and the edge set $\mathcal{E}$ can be partitioned as $\mathcal{E} = \mathcal{E_\text{B}\cup\mathcal{E_\text{D}}}$, where $\mathcal{E_\text{B}}$ includes all base edges connecting fixed stops and $\mathcal{E_\text{D}}$ includes all deviation edges that connect flexible stops and fixed stops. Due to the deviation nature of the system, deviate edges do not geometrically align with base edges. An illustration of the overall setting is given in Figure.~2.
\begin{figure}[!h]
    \centering
    \captionsetup{belowskip=-10pt}
        \includegraphics[width=0.45\textwidth]{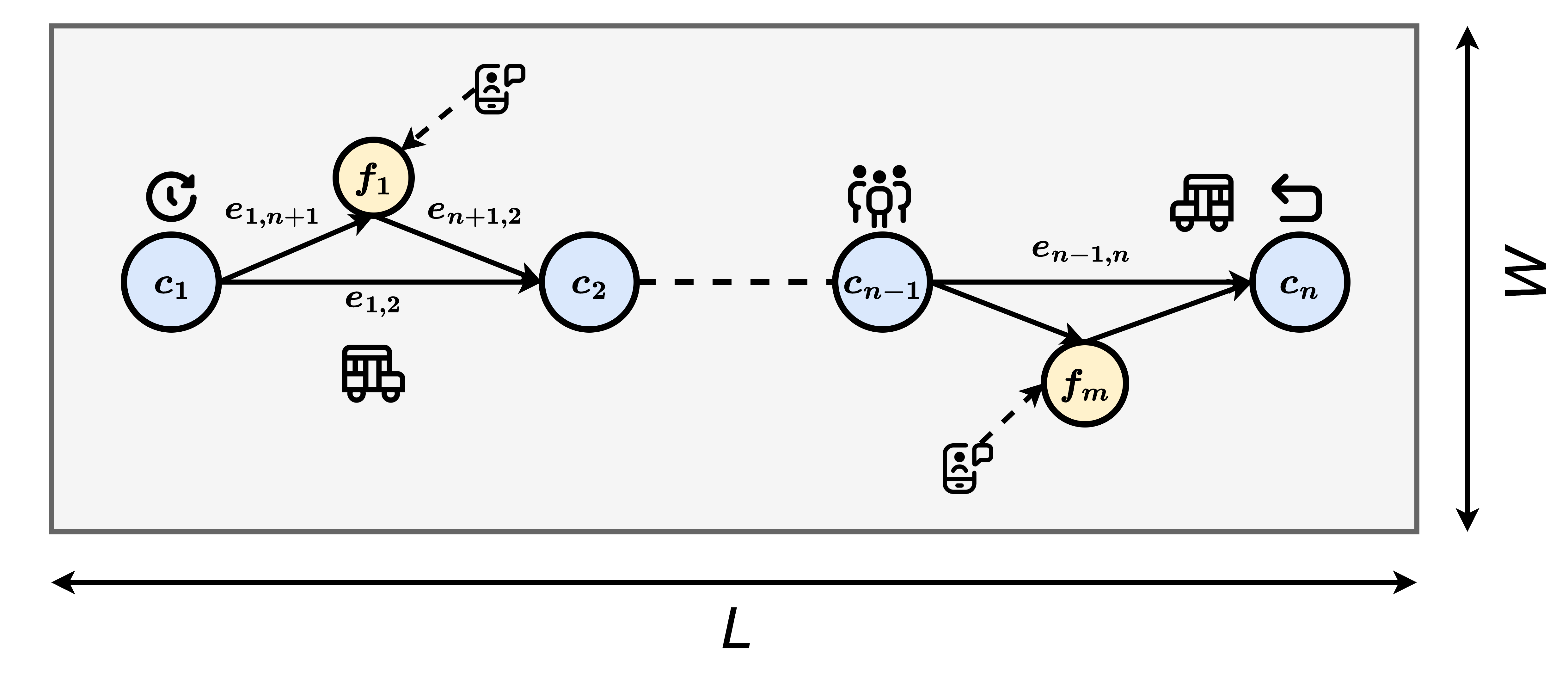}
        \label{fig:2}
    \caption{Flex-Route Transit Operation System Description}
\end{figure}

Each edge $e_{i,j}\in \mathcal{E}$ is assigned a travel time denoted by $T_{\{i,j\}}$. For base edges $e_{i,j}\in\mathcal{E}_\text{B}$, the travel time is a random variable with a normal distribution to model the stochastic traffic conditions, i.e., $T_{\{i,j\}}\thicksim \mathcal{N}(\mu_T,\sigma_T^2)$, where $\mu_T$ and $\sigma_T$ are chosen to represent different congestion levels. For deviation edges $e_{i,j}\in\mathcal{E}_\text{D}$, travel time $T_{\{i,j\}}$ is set to be a constant $T_{\text{D}}$. Then, the travel times are defined as $T = \{T_{\{i,j\}}|~(i,j)\in \mathcal{E}\}$. 

The FRT service uses $K$ homogeneous vehicles, each with a capacity of $Q_c$. Vehicles are dispatched sequentially from stop $c_1$ at intervals of $h$ minutes. If all vehicles have been dispatched and none have returned, the system must wait until a vehicle reaches terminal $c_n$ and returns to the origin stop $c_1$ to initiate the next departure. Passengers comprise (i) regular riders arriving at fixed stops $\mathcal{C}$, and (ii) on-demand riders requesting pick-up at flexible stops $\mathcal{F}$. On-demand passengers make a request in advance or just walk to an on-demand stop. Requests are not proactively rejected, but expire if waiting exceeds a threshold of $W_\text{fixed}$ at $\mathcal{C}$ or $W_\text{flex}$ at $\mathcal{F}$. Passengers with pre-booked deviation requests may not show up with probability $\beta_{1}$. The number of passengers waiting at all stops at time $t$ is represented by the set $\boldsymbol{Q^p_t} = \{q_v(t)|~\forall v \in \mathcal{V}\}$. 

At any time $t$, vehicles are either dwelling at a stop or traversing a directed edge. A binary \emph{occupancy matrix} $Q^\text{o}_t\in\{0,1\}^{N\times N}$ is defined to track vehicles' locations at each time $t$. For all stops $v_i \in \mathcal{V}$:
\[
Q^\text{o}_t(i,i)=
\begin{cases}
1&\text{if stop } v_i \text{ is occupied},\\
0&\text{otherwise},
\end{cases}
\]
And for all $e_{i,j} \in \mathcal{E}$
\[
Q^\text{o}_t(i,j)=
\begin{cases}
1&\text{if edge } e_{i,j}\text{ is occupied},\\
0&\text{otherwise}
\end{cases}
\]

Vehicles make pick-up decisions only upon arrival at control stops, which is a subset of $\mathcal{C}$, to decide whether to proceed along the \emph{base} edge
or detour via the corresponding \emph{deviation} stop.
The objective is to select deviations to serve more requests.
\subsection{Real-time Operation Control}
Since we treat FRT operation control from a reinforcement learning (RL) perspective, we establish it within the partially observed Markov decision process (POMDP) framework. A POMDP can be described as a tuple of $\langle S, A, \prob, R, \Omega, O, \gamma\rangle$. $S$ indicates the set of states, $A$ refers to the action set for the agent, $R$ is the reward, $\Omega$ is the set of observations for an agent, $\prob$ and $O$ are state transition probabilities and observation probabilities, respectively, $\gamma$ is the discount factor. In the context of FRT real-time operation control, we treat the whole system as the control agent, and adaptively control the deviation decisions at each control point. The travel time matrix $T$, passenger distribution $Q^p_t$, vehicle location $Q^o_t$, and the number of idle vehicles at the starting point $c_1$, compose the agent observation $\Omega$. Action $A$ consists of deviation decisions at each control point. The action is updated based on a fixed resolution and vehicle arrival at the fixed stops. The discount factor $\gamma=1$; the instantiated reward $r_t$ is set to be the served passengers at each timestep. Let $\pi(a|s)$ be the probability (density) of taking action $a$ in state $s$ when using policy $\pi$. It is denoted as $\pi(\boldsymbol{\theta})$ if it can be parametrized by vector $\boldsymbol{\theta}$. The objective is to find the policy $\pi(\boldsymbol{\theta})$, which maximizes the cumulative reward.

\section{The Hi-CoLA Framework}
This section presents the proposed Hi-CoLA framework for reliable and efficient policy improvement for the real-time control of FRT systems. As background, we first introduce the high-confidence off-policy evaluation (HCOPE)\cite{Thomas2015HCOPE}, which provides the labels for the Hi-CoLA network pre-training. Then, we present the policy optimization phase that achieves monotonic policy performance improvement under a user-defined confidence guarantee.
\begin{figure*}[!bth]
    \centering
    \captionsetup{belowskip=-15pt}
    \includegraphics[width=1\textwidth]{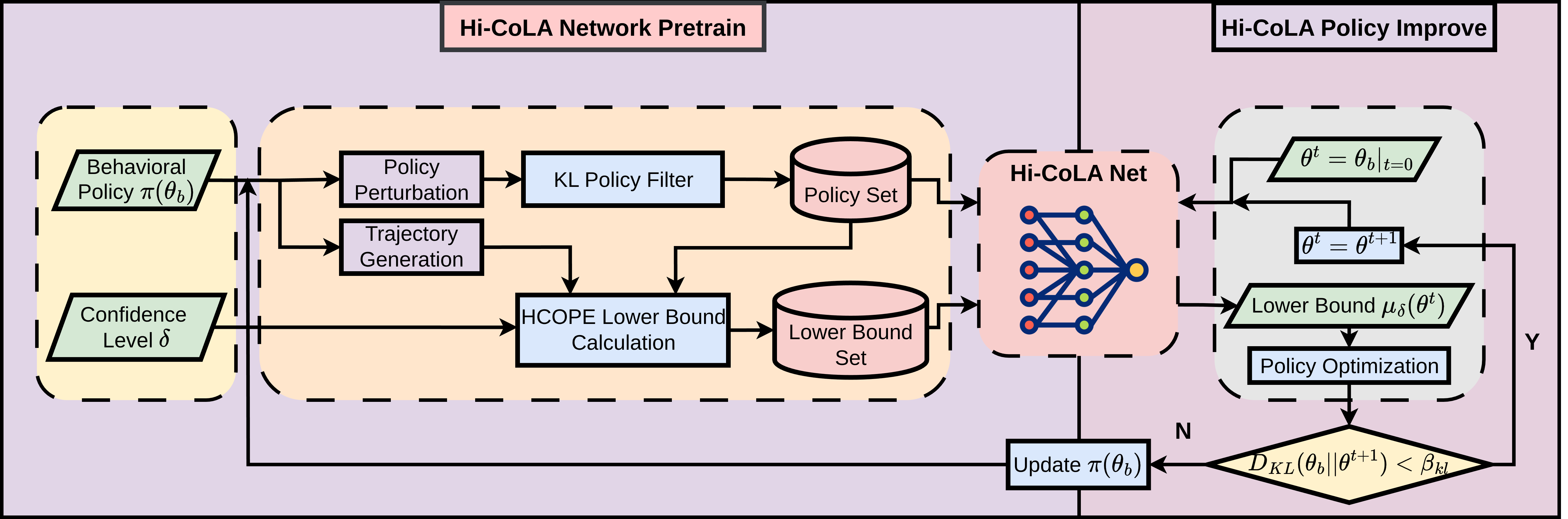}
    \caption{Hi-CoLA framework and the abstract flow process. The framework has
two phases: Hi-CoLA network pre-training and policy improvement. In
pre-training, the behavioral policy $\pi(\theta_b)$ perturbs neighborhood
candidate policies and generates trajectories for confidence lower bound
calculation. The perturbed policies pass through a KL filter, and the
qualified ones serve as training features; their confidence lower bounds,
computed via HCOPE, serve as training labels for the Hi-CoLA Net. The trained
network is then deployed to improve $\pi(\theta_b)$: starting from
$\theta^{t}=\theta_b$, it predicts $\hat{\mu}_\delta(\theta^{t})$, which
gradient-based policy optimization maximizes. Updates iterate until leaving
the KL trust region, after which the behavioral policy is updated and the
cycle repeats for continual improvement.}
    \label{fig:Framework}
\end{figure*}
\subsection{Confidence Guaranteed Policy Performance Estimation}
The operating process of FRT can be represented by a trajectory of length $H$, which is an ordered set of states, actions, and rewards as defined previously:
\[
\tau = \{s_1^{\tau}, a_1^{\tau}, r_1^{\tau}, s_2^{\tau}, a_2^{\tau}, r_2^{\tau}, \ldots, s_H^{\tau}, a_H^{\tau}, r_H^{\tau}\}.
\]
The normalized and discounted return of a trajectory is defined as:
\begin{equation}
    R(\tau) = \frac{\left(\sum_{t=1}^{H} \gamma^{t-1} r_t^{\tau}\right) - R_-}{R_+ - R_-} \in [0,1]
    \label{eq:norm}
\end{equation}
where $R_-$ and $R_+$ are the theoretical lower and upper bounds on the cumulative return for each trajectory in the environment. The bounded performance of the policy $\pi$ parameterized by $\boldsymbol{\theta}$ is denoted by:  
\[
\rho(\pi(\boldsymbol{\theta})) = \mathbb{E}[R(\tau)|\pi(\boldsymbol{\theta})]
\]
A behavioral policy $\pi(\boldsymbol{\theta_b})$ is used to generate a set of trajectories and store them into $\mathcal{D} := \{\tau_i | i \in \{1, \ldots, n_{\tau}\}\}$. With the behavioral policy $\pi(\boldsymbol{\theta_b})$ and the collected trajectories, we can get the unbiased estimates $\hat{\rho}(\boldsymbol{\theta_e},\tau_i,\boldsymbol{\theta_b})$ of the performance of any given evaluation policy $\pi(\boldsymbol{\theta_e})$ using the importance sampling method in Eq.~\ref{eq:2}.
\begin{equation}
    \hat{\rho}(\boldsymbol{\theta_e},\tau_i,\boldsymbol{\theta_b}) = R(\tau_i)\frac{Pr(\tau_i|\boldsymbol{\theta_e})}{Pr(\tau_i|\boldsymbol{\theta_b})} = R(\tau_i)\prod^{H}_{t=1}\frac{\pi(a_t|s_t,\boldsymbol{\theta_e})}{\pi(a_t|s_t,\boldsymbol{\theta_b})}
\label{eq:2}
\end{equation}

After getting the estimates across $\mathcal{D}$, we use the HCOPE method proposed by \cite{Thomas2015HCOPE} to accurately approximate the confidence lower-bound of the performance of $\pi(\boldsymbol{\theta_e})$, which leverages a special form of concentration inequality theorem. Let $\hat\mu_\delta{(\theta_e)}$ be the estimation of confidence lower bound of $(1-\delta)$, i.e. $E(\{\hat{\rho}(\theta_e,\tau_i,\theta_b)|~\forall \tau_i \in \mathcal{D}\})\geq\mu$ has probability at least $(1-\delta)$. The predictor is
\begin{align}
  \hat\mu_\delta{(\boldsymbol{\theta_e)}}
  &= \frac{1}{|\mathcal{D}|} \sum_{i=1}^{|\mathcal{D}|} M_i
    - \frac{7\kappa \ln(2/\delta)}{3(|\mathcal{D}|-1)} \label{eq:hcope} \\   %
  &\quad -
    \sqrt{
      \frac{2\ln(2/\delta)}{|\mathcal{D}|^2(|\mathcal{D}|-1)}
      \left[
        |\mathcal{D}| \sum_{i=1}^{|\mathcal{D}|} M_i^2
        - \left(\sum_{i=1}^{|\mathcal{D}|} M_i\right)^{2}
      \right]
    } \nonumber                                                              %
\end{align}
where $M_i$ is shorthand for $\min\{\hat{\rho}(\boldsymbol{\theta_e},\tau_i,\boldsymbol{\theta_b}),\,\kappa\}$.

To obtain a tight estimator $\hat\mu_\delta{(\boldsymbol{\theta_e)}}$ of the $(1-\delta)$ confidence lower bound, a subset of the data is used to choose the optimal $\kappa^\star$ using the bisection method.
\[
\kappa^\star \in \arg\max_{\kappa \in [1,\infty)} \hat\mu_\delta{(\boldsymbol{\theta_e)}},
\]

\subsection{Hi-CoLA based Policy Improvement}
The previous subsection discussed the background of confidence-integrated policy performance evaluation. In this subsection, we extend the approach to policy improvement by introducing Hi-CoLA. It is designed to deliver performance-guaranteed policy improvement for the real-time control of FRT operations. Hi-CoLA integrates HCOPE into the policy training process by using a differentiable surrogate, namely the Hi-CoLA network, which enables gradient-based policy optimization. The overall policy training scheme is illustrated in Figure 3 and includes two key components: Hi-CoLA network training and gradient-based policy improvement.
\subsubsection{Hi-CoLA Network Training} 
Although HCOPE provides a tight confidence lower bound, it is computationally expensive, and since trajectories with $\hat{\rho}_i \geq \kappa^*$ have zero gradient, the direction of improvement is flattened, making it impractical for iterative policy updates. In this work, we treat the modeling process from policy parameters $\theta$ to the corresponding performance confidence lower-bound $\hat\mu_\delta{(\boldsymbol{\theta})}$ as a machine learning problem, and use the Hi-CoLA network, a multi-layer perceptron (MLP), i.e., $\hat\mu_\delta{(\boldsymbol{\theta)}} = mlp (\boldsymbol{\theta})$. This differentiable proxy amortizes the computational cost of HCOPE and enables backpropagation-based gradient optimization in a trust region.

To generate training data for the Hi-CoLA network, we first execute a behavioral policy $\pi{\boldsymbol{(\theta_b)}}$ that satisfies the basic service requirements of the system and collect its trajectories in the real world in a dataset $\mathcal{D}$ for confidence bound calculation. Around this behavioral policy, we create a set of candidate policies $\pi(\boldsymbol{\theta_p})$ by applying parameter-scaled perturbations with Eq.~\ref{eq:pertub}, where $c_p$ is a constant to adjust the size of the perturbation space, and $\theta^{(i)}$ refers to the indexed parameter in the respective $\boldsymbol{\theta}$. 
\begin{equation}
    {\theta_p^{(k)}} = \theta_b^{(k)} + \epsilon_k, \text{ where } \epsilon_k \thicksim \mathcal{N}(0,c_p\cdot |\theta_b^{(k)}|)
    \label{eq:pertub}
\end{equation}

To ensure that the importance sampling remains stable, we filter the candidates using a KL-divergence threshold $\beta_{kl}$, retaining only those that stay sufficiently close to the behavioral policy. The calculation method is listed below, where  $\hat{\mathcal{S}} \subseteq \mathcal{S}$ is the sampled set of states:
\begin{equation}
\bar{D}_{\mathrm{KL}}(\pi_b \,\|\, \pi_p)
= \frac{1}{|\hat{\mathcal{S}}|}
\sum_{s \in \hat{\mathcal{S}}}
\sum_{a \in \mathcal{A}}
\pi_b(a \mid s)
\log\!\left(
\frac{\pi_b(a \mid s)}{\pi_p(a \mid s)}
\right)
\label{eq:kl}
\end{equation}
\begin{algorithm}[!htp]
{
\caption{Training of Hi-CoLA Network}
\label{alg:CoLA PT}

\LinesNumbered
\KwIn{Behavioral policy $\pi({\boldsymbol{\theta_{b}}})$; Confidence level $\delta$; \newline $\#$ of Trajectories $N_\tau$; $\#$ of policies $N_\pi$; }
\KwOut{Hi-CoLA Network $\hat\mu_\delta{(\boldsymbol{\theta})} = mlp (\boldsymbol{\theta})$}
\kwInit{Hi-CoLA Network: $\hat\mu_\delta{(\boldsymbol{\theta})} = mlp (\boldsymbol{\theta})$; Trajectory Dataset $\mathcal{D} = \emptyset$; Policy Set $\mathcal{D_\theta} = \emptyset$; Lower-Bound Set $\mathcal{D_{\mu}} = \emptyset$; $i,j = 0$   
    }
    
\BlankLine

\While{$i\leq N_\tau$}{
Generate trajectory $\tau_i$ using $\pi({\theta_{b}})$\;
$\tau_i = \{s^{(i)}_1,a^{(i)}_1,r^{(i)}_1,\ldots,s^{(i)}_H,a^{(i)}_H,r^{(i)}_H\}$\;
Store $\tau_i$ in $\mathcal{D}$\;
$i\xleftarrow{}i+1$\;
}

\While{$j\leq N_{\pi}$}{
\ForEach{$k\in\{1,2,\ldots,|\boldsymbol{\theta_b}|\}$}{
 $ {\theta_j^{(k)}} = \theta_b^{(k)} + \epsilon_k, ~ \epsilon_k \thicksim \mathcal{N}(0,c_p\cdot |\theta_b^{(k)}|)$\;
 }
\If{$D_\text{KL}(\boldsymbol{\theta_b} ||\boldsymbol{\theta_j})< \beta_{kl}$}
{
 Store $\boldsymbol{\theta_j}$\;
 Calculate $\hat\mu_\delta{(\boldsymbol{\theta_j})}$ using Eq.~\ref{eq:hcope} with $\mathcal{D}$\;
 Store $\hat\mu_\delta{(\boldsymbol{\theta_j})}$\;
 $j\xleftarrow{}j+1$
 
}
}
Compile $\mathcal{D}_\text{train} =  \{(\theta_1, \mu_1), (\theta_2, \mu_2), \dots, (\theta_{N_\pi}, \mu _{N_\pi})\}$\;
Train $\hat\mu_\delta{(\boldsymbol{\theta})} = mlp (\boldsymbol{\theta})$ with $\mathcal{D}_\text{train}$\;
Final.
}

\end{algorithm}

With Eq.~\ref{eq:hcope}, we can calculate the tight confidence lower bound $\hat\mu_\delta{(\theta_p)}$ for each accepted policy $\pi(\boldsymbol{\theta_p})$, and these policy–lower bound pairs form the supervised training data for the Hi-CoLA network. The detailed training process is summarized in Algorithm 1.

\subsubsection {Hi-CoLA-based Policy Improvement} 
Once trained, the Hi-CoLA network acts as a fast and differentiable surrogate for estimating confidence lower bounds near the behavioral policy. Using backpropagation, we compute the gradient of $\hat\mu_\delta{(\boldsymbol{\theta})}$ with respect to $\boldsymbol{\theta}$, \ie~$\partial\hat\mu_\delta{(\boldsymbol{\theta})}/\partial\boldsymbol{\theta}$. Then we can update the policy parameters using the stochastic gradient descent method with the objective $J = (1-\hat{\mu}(\boldsymbol{\theta}))^2$ since its performance is bounded by $1$ (Eq.~\ref{eq:norm}): 

\begin{equation}
    \;\boldsymbol{\theta}_{}^{\,t+1}
= \boldsymbol{\theta}^{\,t}
- \alpha\,\nabla_{\boldsymbol{\theta}}\!\left(1-\hat{\mu_\delta}(\boldsymbol{\theta})\right)^{2}\;
\end{equation}

This iterative process continues until the updated policy drifts beyond the safe KL-divergence threshold $\beta_{kl}$. Upon convergence, the resulting policy achieves improved performance with a certified confidence guarantee. This policy can then be deployed in the field or used as a new baseline behavioral policy for further refinement. The approach preserves the safety and interpretability of HCOPE while substantially reducing computational overhead, enabling more responsive and reliable policy optimization.

\section{Experiments and Analysis}
In this section, we benchmark the Hi-CoLA performance for FRT real-time operation control with state-of-the-art safe and offline RL algorithms. For the state-of-the-art, we choose (1) Implicit Q-learning (IQL)\cite{2022IQL}; (2) Conservative Q-learning (CQL)\cite{Kumar2020CQL}; (3) TD3 with behavior cloning\cite{Td3+bc}; (4) Daedalus\cite{Thomas2015HighConfidencePI}. IQL, CQL, and TD3+BC are offline RL algorithms, and Daedalus is a safe RL algorithm developed to improve policy with confidence.
The main hyperparameters for selected algorithms, as well as Hi-CoLA, are configured as follows: (1) Hi-CoLA: $\alpha =\num{5e-5}$, $\delta = 0.1$; (2) IQL: $\alpha =\num{1e-4}$; (3) CQL: $\alpha =\num{2e-4}$; (4) TD3+BC: $\alpha =\num{2e-4}$; (5) Daedalus: $\delta=0.1$. All algorithms are implemented in Python 3.9 with publicly available source code \footnote{Source code at \url{https://github.com/kevinkang1125/Hi_CoLA_FRT}.}, and the experiments are conducted on a server equipped with a 16-core, 32-thread CPU, 64 GB RAM, NVIDIA GTX Titan XP GPU, running the 64-bit Ubuntu system.

In the following subsections, we first present a simple yet illustrative scenario to showcase Hi-CoLA's capability of guaranteeing monotonic policy improvement, then benchmark the performance of the Hi-CoLA algorithm with the state of the art in terms of performance distribution and CVaR, and finally examine Hi-CoLA's robustness to the quality of the initial behavioral policy. 
\subsection{Performance Improvement Guarantee}
This subsection demonstrates the performance improvement guarantee of the Hi-CoLA training framework in a simplified FRT environment with $3$ fixed stops and $2$ flexible stops in between. The topology of the route and service direction are shown in Fig. 4(a). 
\begin{figure}[!thbp]
    \centering
    \subfigure[Route Map]{%
        \includegraphics[width=0.11\textwidth]{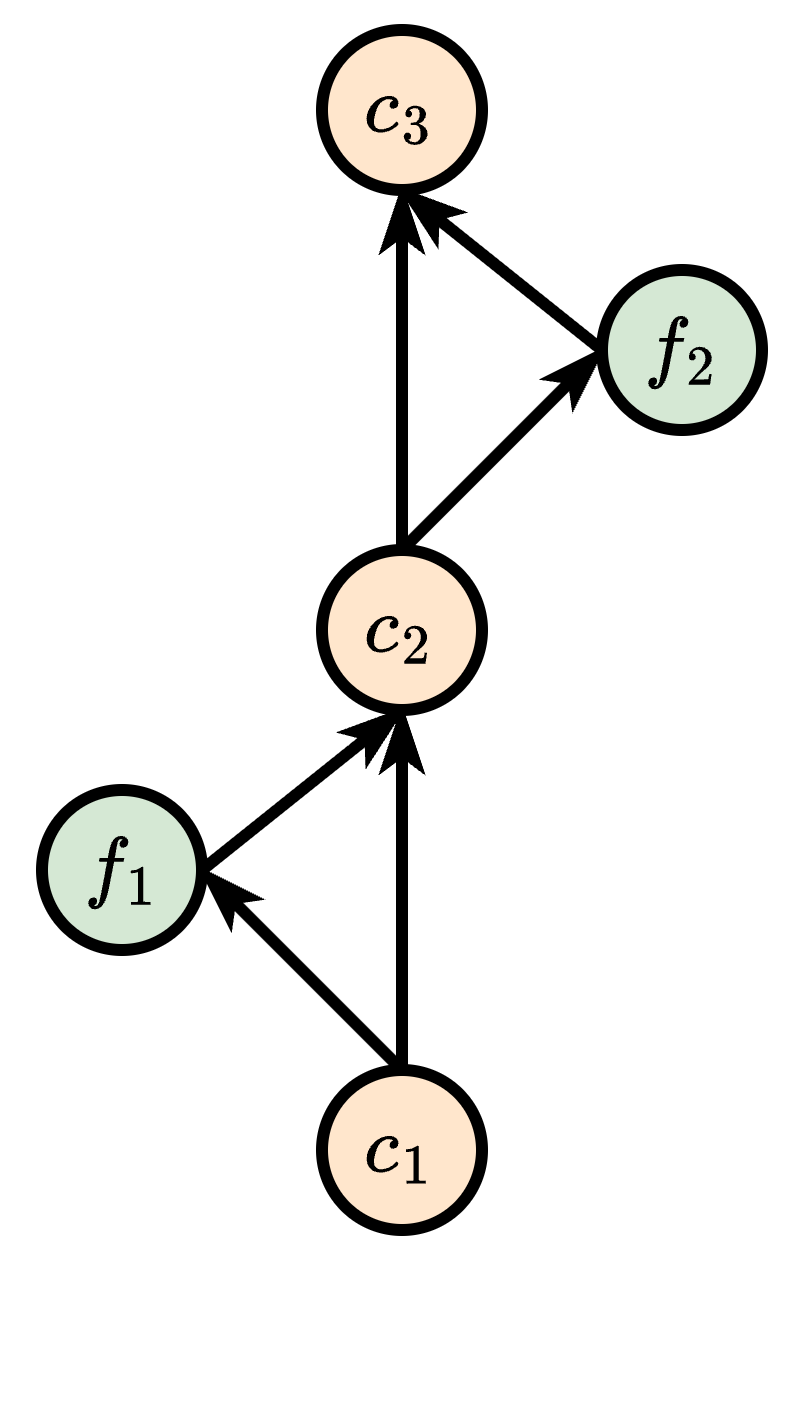}
        \label{fig:simple-scene}}
    \subfigure[Training Process and Policy Improvement]{%
        \includegraphics[width=0.34\textwidth]{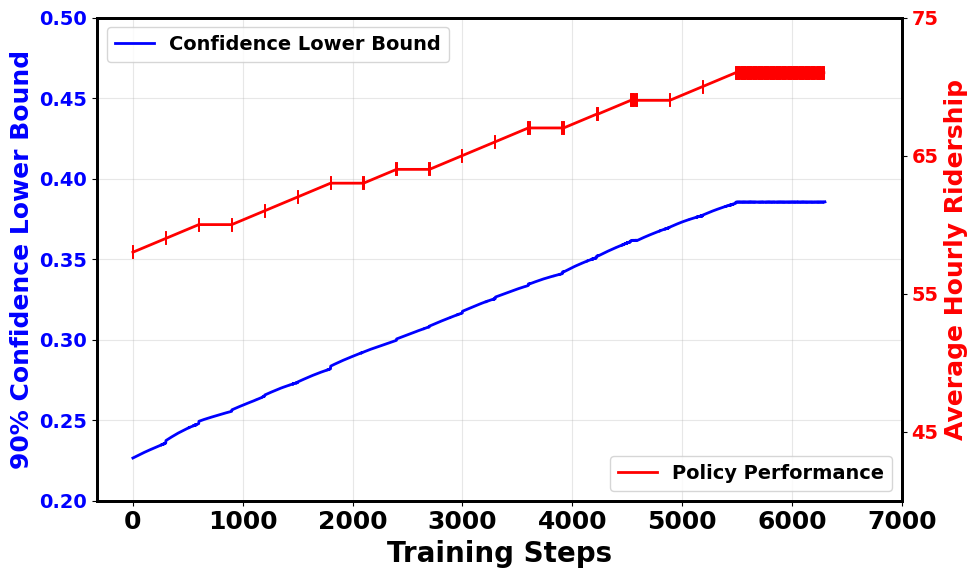}
        \label{fig:learning curve}}
    \caption{Training process}
    \label{fig:4}
\end{figure}
For each trip, the vehicles start at node $c_1$ and make the deviation decisions when they arrive at fixed stops $c_1$ and $c_2$ (decision nodes). Passenger arrivals are sampled from a Poisson distribution with an average arrival rate of $18$ pax/hour for fixed stops and $10$ pax/hour for flexible stops. The maximum waiting time for each passenger is $30$ minutes, and vehicles are dispatched every $12$ minutes.
Initially, the vehicles are controlled by a rule-based behavioral policy, which has a $0.6$ probability to deviate if the passenger requests at $ f_i$ are more than half of the passengers waiting at $c_{i+1}$; otherwise, the deviation probability is $0.2$. Then, we collect $6000$ pairs of observations and corresponding actions to train the policy network imitating this rule-based policy for parameterization purposes. The parameterized policy is used as the initial behavioral policy. Fig. 4(b) shows the evolution of the $90\%$ confidence lower bound at each step and policy performance in terms of the average number of passengers served per hour after each training cycle. From the plot, we can see that as training progresses, the performance of the trained policy monotonically improves, and the updated behavioral policies consistently outperform their predecessors.
\subsection{Benchmarking with State-of-the-Art Approaches}
This subsection compares the Hi-CoLA-based FRT real-time operation policy with state-of-the-art methods in terms of hourly ridership distribution and CVaR. The evaluation environment includes $5$ fixed stops and $4$ flexible stops. To model the dynamics of a real transportation system, the travel time between fixed stops is randomly drawn from a normal distribution and updated at each timestamp. Passengers waiting at flexible stops have a $0.2$ probability of canceling the ride, and the maximum waiting time (after which the passenger leaves the system) is $15$ minutes. The passenger arrivals are sampled from a Poisson distribution with an average arrival rate of $9.6$ pax/hour for fixed stops and $7.5$ pax/hour for flexible stops. All of the approaches used for benchmarking adopt the same initial behavioral policy, which is the same as the one we used in the previous subsection. 
\begin{figure}[!ht]
    \centering
    \captionsetup{belowskip=-10pt}
        \includegraphics[width=0.48\textwidth]{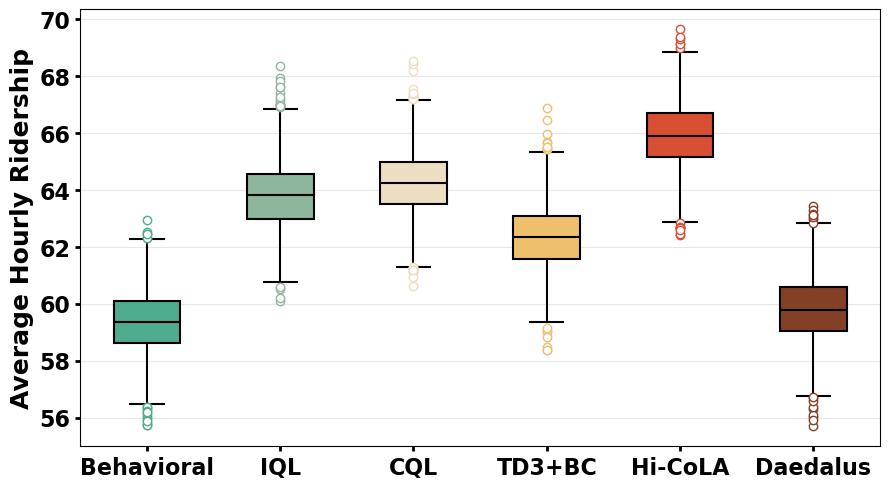}
        \label{fig:5}
    \caption{Performance Distribution Comparison}

\end{figure}

\begin{table}[H]
\centering
\caption{Hourly Ridership CVaR Performance}
\label{tab:cvar_comparison}
\renewcommand{\arraystretch}{1.15}
\setlength{\tabcolsep}{5pt}
\begin{tabular}{ccccc}
\hline
Method & CVaR$_{5\%}$ & CVaR$_{10\%}$ & CVaR$_{5\%}$ Improve & CVaR$_{10\%}$ Improve \\
\hline
Behavioral & 57.07 & 57.41 & 0.00\% & 0.00\% \\
IQL      & 61.39 & 61.75 & 7.57\% & 7.56\% \\
CQL      & 61.95 & 62.28 & 8.55\% & 8.48\% \\
TD3+BC   & 59.97 & 60.35 & 5.08\% & 5.12\% \\
Hi-CoLA  & \textbf{63.52} & \textbf{63.88} & \textbf{11.30\%} & \textbf{11.27\%} \\
Daedalus & 57.51 & 57.86 & 0.77\% & 0.78\% \\
\hline
\end{tabular}
\end{table}
The results in Fig. 5 and Table II demonstrate that Hi-CoLA consistently outperforms all baselines in both average performance and tail-risk metrics. From the distributional perspective, Hi-CoLA achieves the highest median hourly ridership, while also exhibiting a tighter spread compared to other methods, indicating improved stability. More importantly, in terms of risk measurement, Hi-CoLA achieves the best CVaR performance by reaching 63.52 and 63.88 for CVaR$_{5\%}$ and CVaR$_{10\%}$, respectively. The corresponding improvement over the behavioral policy outperforms the performance of the offline RL baselines such as IQL, CQL, TD3+BC, and Daedalus. These results highlight that Hi-CoLA not only improves average performance but also substantially enhances worst-case outcomes.
\subsection{Robustness to the Quality of Initial Behavioral Policy}
Offline reinforcement learning heavily relies on the data collected by the behavioral policy, and thus the quality of the initial behavioral policy plays a vital role in the training process and may impact the performance of the final policy. To evaluate and compare Hi-CoLA's robustness to the quality of the behavioral policy, we repeat the evaluation environment of Section V-B, but replace the rule-based behavioral policy with one that deviates with an independent probability of $0.5$ at every control point. All the other environment settings are the same as in Section V-B, and CVaR$_{5\%}$ and CVaR$_{10\%}$ with the new initial behavioral policy and comparison with their performance in Table II are reported in Table III.
\begin{table}[H]
\centering
\caption{Hourly Ridership CVaR Performance with Random Deviation Policy and Comparison with Table II}
\label{tab:cvar_comparison_postchange}
\footnotesize
\setlength{\tabcolsep}{5pt}
\begin{tabular}{ccccc}
\hline
Method & CVaR$_{5\%}$ & CVaR$_{10\%}$ & $\Delta$ CVaR$_{5\%}$& $\Delta$ CVaR$_{10\%}$\\
\hline
Behavioral & 53.56 & 53.92 & -6.15\% & -6.08\% \\
IQL      & 61.13 & 61.48 & -0.42\% & -0.43\% \\
CQL      & 55.72 & 56.08 & -10.06\% & -9.96\% \\
TD3+BC   & 53.78 & 54.17 & -10.32\% & -10.24\% \\
Hi-CoLA  & \textbf{63.41} & \textbf{63.79} & \textbf{-0.17\%} & \textbf{-0.14\%} \\
Daedalus & 56.29 & 56.46 & -2.12\% & -2.42\% \\
\hline
\end{tabular}
\end{table}
Table III shows that CQL and TD3+BC show the largest reductions in their tail-risk performance when trained from a poorly performing behavioral policy, while Hi-CoLA and IQL's CVaR degrade only marginally by comparison, indicating their robustness to the behavioral policy performance. The performance of Hi-CoLA is consistent with its design: its trust region constraint and confidence-lower-bound objective jointly contribute to guaranteed performance improvement in policy updates. Notably, the robust performance of IQL is consistent with its mechanism as well. Although the random initial behavioral policy performs poorly in terms of served passengers, its strong stochasticity enables it to have a richer coverage of the action space and helps the learning of IQL. In contrast, other methods either try to mimic the behavioral policy or learn from states associated with low performance, resulting in poorer final performance.
\section{Case Study}

In this section, we apply the Hi-CoLA-based FRT real-time operation strategy to a real-world case. We evaluate its performance with a realistic, detailed simulation model SUMO\cite{SUMO}. 

\begin{figure}[!hbp]
    \centering
    \captionsetup{belowskip=0pt}
    \subfigure[Bus Deviation Map]{%
        \includegraphics[width=0.245\textwidth]{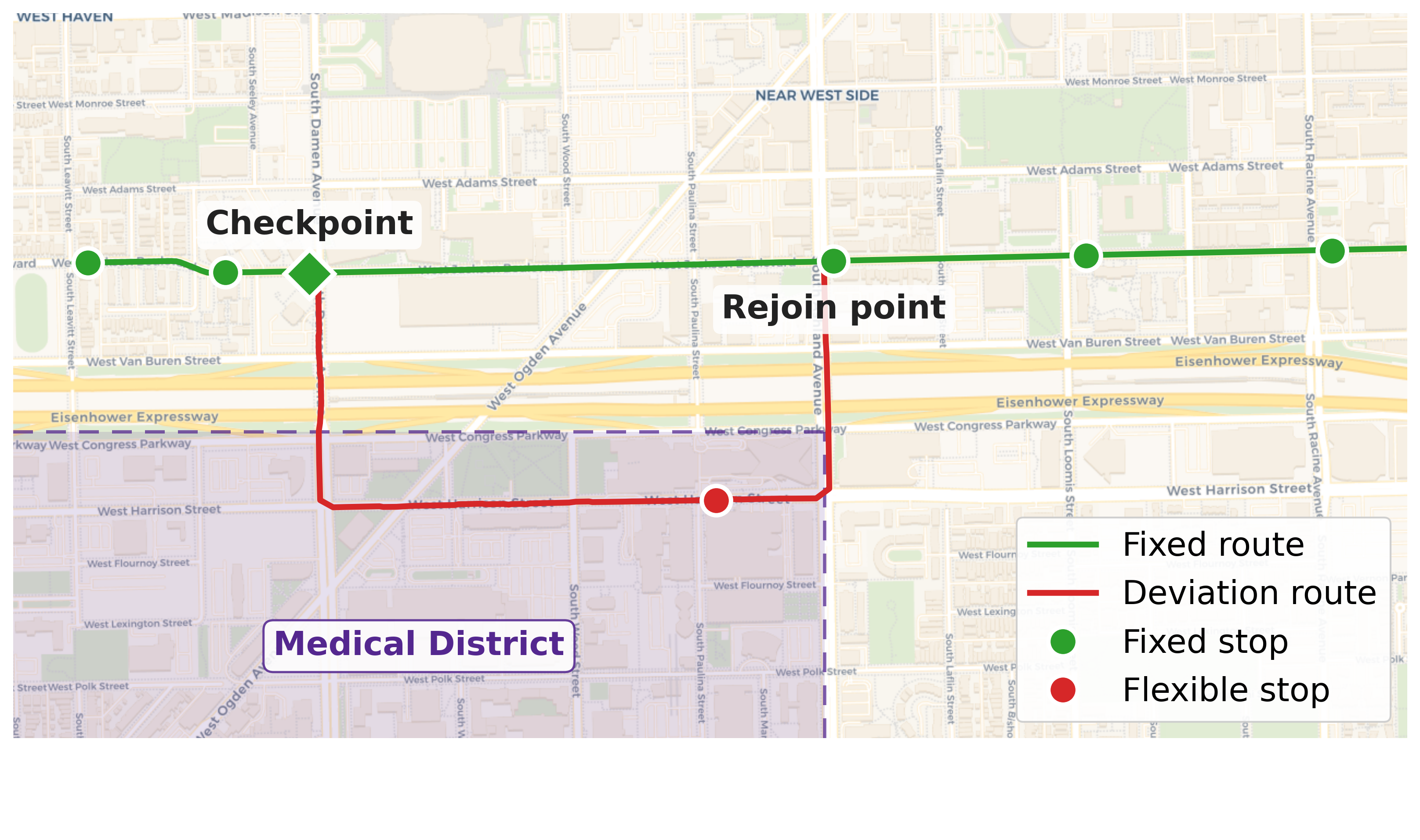}
        \label{fig:CTA}}
    \subfigure[Learning Curve]{%
        \includegraphics[width=0.222\textwidth]{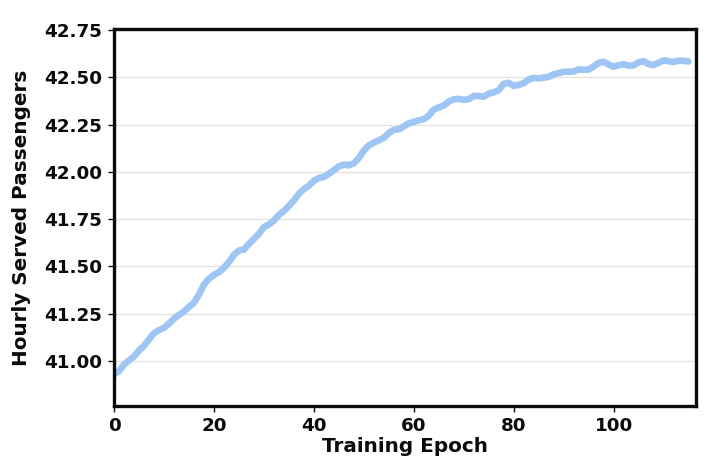}
        \label{fig:lr}}
    \caption{Case Study Illustration}
    
    \label{fig:6}
\end{figure}
The bus service of interest passes through a major medical district in Chicago. During weekdays, the route follows its primary street and switches to a deviation route on a parallel street during weekday late evenings and throughout weekends to serve a hospital area. The deviation maintains connectivity when many other transit options may be reduced. The route is shown in Fig. 6(a).

To improve accessibility to the medical area during weekday daytime hours, we adapted the fixed-schedule deviation service into an on-demand one and deployed Hi-CoLA to train a policy that makes real-time deviation decisions. Service headway is assumed to be 15 minutes, and the behavioral policy keeps the bus on the fixed route with probability $0.99$ (mimicking the current weekday daytime service). Passenger arrivals at each stop follow a Poisson process with rates of 7.5 pax/hour at fixed stops and 4 pax/hour at the flexible stop. Passengers at fixed and flexible stops leave the system once their waiting time exceeds 15 minutes and 25 minutes, respectively, and the resulting passenger loss is penalized in the reward. Travel time depends on traffic conditions, and the deviation route takes approximately 6 minutes longer than the fixed route on average. An inefficient deviation serves few or no additional passengers while delaying the bus's arrival at downstream fixed stops and causing passenger loss. 

The learning curve is shown in Fig. 6(b). Under the Hi-CoLA training framework, performance improves steadily over training. The updated policy produces efficient deviations, serving, as expected, more passengers.
\section{CONCLUSION}
This paper presented Hi-CoLA, a High-Confidence Lower Bound Approximation-based reinforcement learning framework for safe policy improvement in real-time control of flex-route transit (FRT) systems. By integrating high-confidence off-policy evaluation into a differentiable surrogate network, Hi-CoLA enables gradient-based policy updates that are certified, with user-specified confidence, addressing a central barrier in deploying RL in safety- and cost-critical operational settings.
We demonstrated that Hi-CoLA achieves monotonic policy improvement and outperforms state-of-the-art offline and safe RL baselines in both average and tail-risk (CVaR) performance. A case study converting a scheduled deviation segment of an actual bus route into an on-demand service, simulated in SUMO, further illustrated the framework's applicability to actual transit service.

In the future, we are keen on improving the Hi-CoLA framework by extending Hi-CoLA to multi-agent coordination settings, integrating it with large-scale operational data pipelines, and applying it to other domains where performance guarantees are essential.

\clearpage

\bibliographystyle{./Bibliography/IEEEtran}
\bibliography{./Bibliography/IEEEexample}
\clearpage

\end{document}